\documentclass[5p,times]{elsarticle}

\PassOptionsToPackage{hyphens}{url}\usepackage{hyperref}
\usepackage{amsmath,amssymb,amsfonts}
\usepackage{algorithmic}
\usepackage{graphicx}
\usepackage{textcomp}
\usepackage{xcolor}
\usepackage{caption}
\usepackage{subcaption}
\usepackage{float}
\usepackage{amsmath}
\usepackage{booktabs}

\usepackage{tikz}
\usepackage{filecontents}
\usepackage[labelfont=bf]{caption}
\usepackage{breakcites}
\usepackage{lineno}
\usepackage[linesnumbered,ruled,vlined,onelanguage]{algorithm2e}
\usepackage{hyphenat}
\usepackage{subcaption}
\usepackage{etoolbox}
\usepackage{verbatim}
\usepackage{pdfpages}

\begin{document}
\begin{sloppypar}

\begin{frontmatter}

% Siamak: my sugesstion:
\title{Anomal-E: A Self-Supervised Network Intrusion Detection System\\ based on Graph Neural Networks}

% \title{Anomal-E: A Self-Supervised Graph Neural Network based Network Intrusion Detection System}

\cortext[cor]{Corresponding author}

\author[1]{Evan Caville\corref{cor}\fnref{fn1}}
\ead{e.caville@uq.net.au}

\author[1]{Wai Weng Lo\corref{cor}\fnref{fn1}}
\ead{w.w.lo@uq.net.au}

\author[1]{Siamak Layeghy}
\ead{siamak.layeghy@uq.net.au}

\author[1]{Marius Portmann}
\ead{marius@itee.uq.edu.au}

\fntext[fn1]{The first two authors contributed equally to this work.}
\fntext[fn1]{Accepted by Knowledge-Based Systems, Elsevier}

\address[1]{School of ITEE, The University of Queensland, Brisbane, Australia}

\begin{abstract}
This paper investigates graph neural networks (GNNs) applied for self-supervised intrusion and anomaly detection in computer networks. GNNs are a deep learning approach for graph-based data that incorporate graph structures into learning to generalise graph representations and output embeddings. As traffic flows in computer networks naturally exhibit a graph structure, GNNs are a suitable fit in this context. The majority of current implementations of GNN-based network intrusion detection systems (NIDSs) rely on labelled network traffic. This limits the volume and structure of input traffic and restricts the NIDSs' potential to adapt to unseen attacks. These systems also rely on the use of node features, which may reduce the detection accuracy of these systems, as important edge (packet-level) information is not leveraged. To overcome these restrictions, we present Anomal-E, a GNN approach to intrusion and anomaly detection that leverages edge features and a graph topological structure in a self-supervised manner. This approach is, to the best of our knowledge, the first successful and practical approach to network intrusion detection that utilises network flows in a self-supervised, edge-leveraging GNN. Experimental results on two modern benchmark NIDS datasets display a significant improvement when using Anomal-E compared to raw features and other baseline algorithms. This additionally posits the potential Anomal-E has for intrusion detection on real-world network traffic.
\end{abstract}


\end{frontmatter}

\section{Introduction}
The increasing frequency and complexity of attacks on computer networks continues to threaten the security of information within computer systems. To combat this, enterprises utilise network intrusion detection systems (NIDSs) to protect critical infrastructure, data, and networks. These systems commonly identify intrusions through the analysis of raw network traffic, in the form of packet captures or flow-based records (traffic categorised with features detailing statistics about said flow). Traditionally, NIDSs can be broken into two main categories, signature-based and behavioural-based. Signature-based NIDSs utilise a pre-determined set of rules, metrics, or calculations to classify and address network traffic. Behavioural NIDSs, on the other hand, rely on more complex operations, commonly involving machine learning (ML) to identify sophisticated and ever-evolving attacks. These behaviour-specific approaches commonly use supervised learning, which is not always feasible, as network traffic is rarely categorised, especially in the context of zero-day attacks. Furthermore, NIDS approaches have generally not considered the topological patterns of both benign and attack network traffic, which is an important factor in network intrusion detection. Due to this, we strongly argue for the inclusion of network flow topological patterns in NIDS development. This inclusion can be leveraged to detect sophisticated attacks, such as advanced persistent threat (APT) attacks, as overall network graph patterns and lateral movement paths are considered.

Within this field, graph neural networks are a promising recent development in deep learning, as they offer the ability to leverage topological patterns in training and testing. These neural networks take advantage of the graph structures through the use of messages passing between nodes and/or edges. This enables the neural network to learn and generalise effectively on graph-based data to output low-dimensional vector representations (embeddings) \cite{wu2020comprehensive}.

In this paper, we propose Anomal-E, a self-supervised GNN-based NIDS. Given the severe lack of real-world and labelled network traffic, it is difficult to train a supervised NIDS capable of detecting real network attacks. Therefore, to mitigate reliance on labelled data, Anomal-E utilises a self-supervised approach to intrusion detection. Computer network traffic is naturally graph-based, as hosts represent nodes, and edges describe the flows or packets sent between them. Therefore, this is well suited for the graph-based, self-supervised network intrusion detection we present, having no need for data labels.

Dominant graph-based learning approaches \cite{hamilton2017inductive, grover2016node2vec} use only node/topological features and rely commonly on a random walk strategies, which can lead to potentially important network flow information (edge information) being lost in the learning process. For NIDSs, edge information is critical in identifying and mitigating network intrusions and anomalies. Anomal-E overcomes this restriction and leverages graph and edge information, rather than node information, in a self-supervised learning manner. Anomal-E consists of two main components. The first component is E-GraphSAGE \cite{lo2021graphsage}. E-GraphSAGE extends the original GraphSAGE \cite{hamilton2017inductive} model to capture edge features and topological patterns in graphs. The second critical component in Anomal-E is the modified deep graph infomax (DGI) method \cite{velickovic2018deep}, which maximises the local mutual edge information between different parts of the input in the latent space for self-supervised learning.

Anomal-E is specifically designed and implemented for the problem of network intrusion detection. The unique combination of its two key components allows for a completely self-supervised approach to the generation of edge embeddings without using any data labels; the generated edge embeddings can then be fed to any traditional anomaly detection algorithm, such as isolation forest (IF) \cite{liu2008isolation}, for network intrusion detection. 

To the best of our knowledge, our approach is the first thoroughly analysed, evaluated, practical, and successful approach to designing self-supervised GNNs for the problem of network intrusion detection.

In summary, the key contributions of this paper are:

\begin{itemize}
\item 
Anomal-E, a self-supervised GNN-based approach for network intrusion detection. This method allows for the incorporation of both edge features and topological patterns to detect attack patterns, without the need for labelled data. 

\item 
Anomal-E is based on E-GraphSAGE, modified DGI, and four different traditional anomaly detection algorithms, namely, PCA-based anomaly detection (principal component analysis) \cite{shyu2003novel}, IF, clustering-based local outlier factor (CBLOF) \cite{he2003discovering}, and histogram-based outlier score (HBOS) \cite{goldstein2012histogram}, for network intrusion detection. Our proposal is the first to utilise self-supervised GNNs for network intrusion detection, to the best of our knowledge.

\item 
We apply Anomal-E on two benchmark NIDS datasets for network intrusion detection. Significant improvements over raw features demonstrate its potential via extensive experimental evaluation. 

\item 
We combine additional graph-based learning methods with anomaly detection algorithms for baseline comparisons. To the best of our knowledge, our work is the first to apply these novel and unique approaches for network attack detection.
\end{itemize}

The rest of the paper is organised as follows. Section~\ref{related} discusses related works, and Section~\ref{Background} provides an overview of the relevant background. Our proposed Anomal-E algorithm and the corresponding NIDS are introduced in Section~\ref{anoml-E}. Experimental evaluation results are presented in Section~\ref{results}. Section \ref{conclusion} uses the experimental results to argue for the practical applications of Anomal-E and discusses future research directions.

 \section{Related Work} \label{related}
 
As our work utilises methods at the forefront of GNNs and applies them to network anomaly detection in a self-supervised context, there is a limited amount of prior work that we know of which is closely related to our research. Therefore, the prior research discussed below relates to our approach in a more broad sense.

\subsection{Graph-Based Approaches for Network Intrusion/Anomaly Detection}

Zhou et al. \cite{zhou2020automating} proposed a GNN model for end-to-end botnet detection. In this approach, the authors consider only the topological structure of the network rather than the features associated with each node and edge. To simulate a realistic network, authentic background traffic was embedded into existing benchmark botnet datasets. To focus on only topological structure, edge features were omitted, and node features were set to constant vector-containing-only ones. This enables the model to aggregate features throughout the graph without any consideration of or effects from unique node attributes. %Please ensure intended meaning is retained.
The authors used this approach to graph learning to formulate a binary classification problem to identify botnets. They found that their proposed model significantly improved botnet detection when compared to results achieved by logistic regression and the existing botnet detection tool, BotGrep \cite{grep}.

\cite{lopez2021network}\cite{Xiao} proposed a traditional graph embedding approach to perform network intrusion detection. They first converted the network flows into graphs based on source and destination IP and port pairings and then applied traditional graph embedding techniques such as DeepWalk (skip-gram) for network intrusion detection. However, a significant limitation of this approach is the application of traditional transductive graph embedding methods \cite{Hamilton2017}, which cannot generalise to any unseen node embeddings, e.g., IP addresses and port numbers, not included during the training phase. This makes the approach unsuitable for most practical NIDS application scenarios, as we cannot expect every IP and port pair to appear in the training data.

\subsection{Alternate Approaches for Network Intrusion/Anomaly \mbox{Detection}}

Layeghy et al. \cite{siamak} investigated the performance and generalisation of ML-based NIDSs by comprehensively evaluating seven different supervised and unsupervised machine learning algorithms on four recently published benchmark NIDS datasets. For the unsupervised learning experiments, IF, one-class support vector machines (oSVM), and stochastic gradient descent one-class support vector machines (SGD-oSVM) were considered. Overall, the results demonstrated that unsupervised anomaly detection algorithms can generalise better than supervised learning methods.

Casas et al. \cite{casas2012unsupervised} proposed an unsupervised NIDS utilising sub-space clustering, DBSCAN, and evidence accumulation (EA4RO). Sub-space clustering enables feature vectors to be projected into k-dimensional sub-spaces, and each sub-space is then grouped by DBSCAN algorithms. Due to DBSCAN's use of density to identify clusters, anomalies are located in the low-density clusters. Finally, EA4RO is applied for calculating the degree of dissimilarity from an outlier to the centroid of the largest cluster, using the Mahalanobis distance as the metric. This algorithm was evaluated using the KDD'99 dataset and achieved 100\% accuracy. However, this dataset is over two decades old and does not reflect the current sophistication of network attacks.

In \cite{syarif2012unsupervised}, the authors study and compare the performances of various anomaly-detection-based methods, including the k-medoids, k-means, expected maximisation (EM), and distance-based algorithms for unsupervised network intrusion detection. Experiments were evaluated using the NSL-KDD dataset. This dataset is an improved version of the KDD'99 dataset, solving redundant problems, as about 78\% of the records are duplicated in the training set and 75\% are duplicated in the test set. Overall, they achieved 78.06\% accuracy in the EM clustering approach, 76.71\% accuracy using k-Medoids, 65.40\% when using improved k-means, and 57.81\% using k-means. However, all algorithms used in the experiments suffer from high false-alarm rates (FAR). The overall rate across the four algorithms was 20\%.  %Please ensure intended meaning is retained.

Leichtnam et al. \cite{leichtnam2020sec2graph} defined a security object graph and applied an autoencoder (AE) model to perform network intrusion detection in an unsupervised manner. The authors evaluated the proposed model's performance on the CICIDS2017 dataset and reported that they can achieve a binary classification performance with an F1-score of 0.94 and classification accuracy of 97.481%.

Monowar H. Bhuyan et al. \cite{bhuyan2012effective} proposed a tree-based sub-space clustering technique to find clusters among intrusion detection data, without using labels. The network data were, however, labelled using a clustering technique based on a TreeCLUS algorithm. This proposed method works faster for detecting intrusions in the numeric mixed categories of network data.

\subsection{Novel Graph-Based Studies}

Similarly to the approach we present, research conducted by Song et al. \cite{SONG2022108274} leveraged self-supervised GNNs to learn and represent graph-based data. However, rather than applying it to NIDSs, the authors applied their method to knowledge tracing (KT). Their proposed model, bi-graph contrastive-learning-based knowledge tracing (Bi-CLKT), aims to learn in a way that captures both local node information and global graph information. To achieve this, node and graph embeddings are used to learn through a positive and negative pair graph training approach. The success of the self-supervised GNN was shown through significant improvements authors were able to gain over the state-of-art models used for KT.

Yang et al. \cite{yang2021variational} posited another self-supervised GNN approach, variational co-embedding learning for attributed network clustering (VCLANC). The authors' method utilises node embeddings in conjunction with their attributes to leverage the local and wider mutual affinities. These graph elements must effectively cluster graph-based data. By doing this, this approach breaches the limitations of other graph-based clustering processes.

The authors of \cite{yang2021interpretable} investigated graph convolutional networks (GCNs) for heterogeneous information networks (HINs). Several key deficiencies hinder the learning ability of GCNs on HINs, restricting their accuracy and efficiency. To overcome this, the authors proposed an interpretable and efficient heterogeneous graph convolutional network (ie-HGCN). Results from experiments on three real heterogeneous network datasets demonstrated the ability the researchers' model has to effectively understand and learn on HINs to overcome limitations seen in prior state-of-the-art approaches in this problem space.

Alshammari et al. \cite{ALSHAMMARI2022100192} investigated parameter-less graph reduction for graph-based clustering methods. By having no parameters during graph reduction, authors' removed the need for dataset-specific parameter tuning, enabling for a more generalised and less manual clustering process. Using this parameter free, graph-based clustering alternative, authors' can competitively challenge other parameter-based approaches.

\section{Background}\label{Background}

Graph representation learning is a fast-growing area of research that can be applied to various applications, including telecommunication and molecular networks. Recently, GNNs have achieved state-of-the-art performance in cyberattack detection, such as network intrusion detection \cite{lo2021graphsage}. 

\subsection{Graph Neural Networks}
    GNNs at a very high level can be described as a deep learning approach for graph-based data, and a recent, highly promising area of ML. GNNs acts as an extension to convolutional neural networks (CNNs) that can generalise to non-euclidean data \cite{wu2020comprehensive, zhou2020graph}. A key feature of GNNs is their ability to use the topological graph structure to their advantage through message passing. For each node in a graph, this means aggregating neighbouring node features to leverage a new representation of the current node that takes into account neighbouring information \cite{xu2018powerful}. The outputs of this process are embeddings \cite{cai2018comprehensive}. For nodes, embeddings are low or n-dimensional vector representations that capture topological and node properties. This process can be performed for edges and graphs in a similar fashion to output edge and graph embeddings. Embeddings are then commonly used for both supervised and unsupervised downstream tasks, e.g., node classification, link prediction, and anomaly detection.

    As computer networks and their traffic flows are naturally able to be represented in graph form, the application of GNNs for network intrusion and anomaly detection is a promising approach for detecting new and advanced network attacks, such as APT attacks. In this context, nodes can be represented by IP addresses, and edges are represented by the communication of packets or flow between nodes \cite{Zhou2018GraphNN}.

\subsection{GraphSAGE}\label{GraphSAGE Algorithm}

\textit{Graph Sample and Aggregate (GraphSAGE)} is one of the baseline graph neural network models proposed by Hamilton et al. \cite{Hamilton2017}. In GraphSAGE, a neighbour sampling technique can be applied to sample a fixed-size subset of node neighbours for neighbour message propagation. The neighbour sampling aims to minimising the size of the computational graph to reduce the space and time complexity of the algorithm.

The GraphSAGE algorithm performs graph convolutional operations on a graph $\mathcal{G}(\mathcal{V}, \mathcal{E})$, where nodes are represented as $\mathcal{V}$, edges as $\mathcal{E}$ and node features as $x_v$.

For specifying the depth of GNNs, a $k$-hop neighbourhood needs to be set. This represents $k$-hop node neighbour information that needs to be aggregated at each iteration. Another important aspect of GNNs is the choice of a differentiable aggregator function $AGG_k, k\in \{1,...,K\}$ for neighbour information aggregation. In GraphSAGE, various aggregation methods can be applied, including mean, pooling, or using different types of neural networks, e.g., long short-term memory (LSTM)~\cite{Hamilton2017}. 

\subsubsection{Node Embedding}
For each node, the GraphSAGE model iteratively aggregates neighbouring node information at $k$-hop depth. At each iteration, a set of the neighbour's nodes can be sampled to reduce the space and time complexity of the algorithm. The representation from the sampled aggregated nodes is used for generating node embeddings.

At the $k$-th layer, the aggregated node neighbour information $\mathbf{h}_{N(v)}^{k}$ at a node $v$, based off the sampled neighbourhood $N(v)$, can be expressed as Equation \ref{eq:general_aggregator} \cite{Hamilton2017}:

\begin{equation}
\label{eq:general_aggregator}
    \mathbf{h}_{\mathcal{N}(v)}^{k} = \text { AGG }_{k}\left(\left\{\mathbf{h}_{u}^{k-1}, \forall {u} \in \mathcal{N}(v)\right\}\right)\
\end{equation}

The node embedding  $\mathbf{h}_{u}^{k-1}$ indicates the node embedding $u$ in the previous layer. These neighbouring node embeddings are aggregated into the embedding of node $v$ at layer $k$.

\begin{figure}[!t]
    \centering
        \includegraphics[width=0.95\columnwidth]{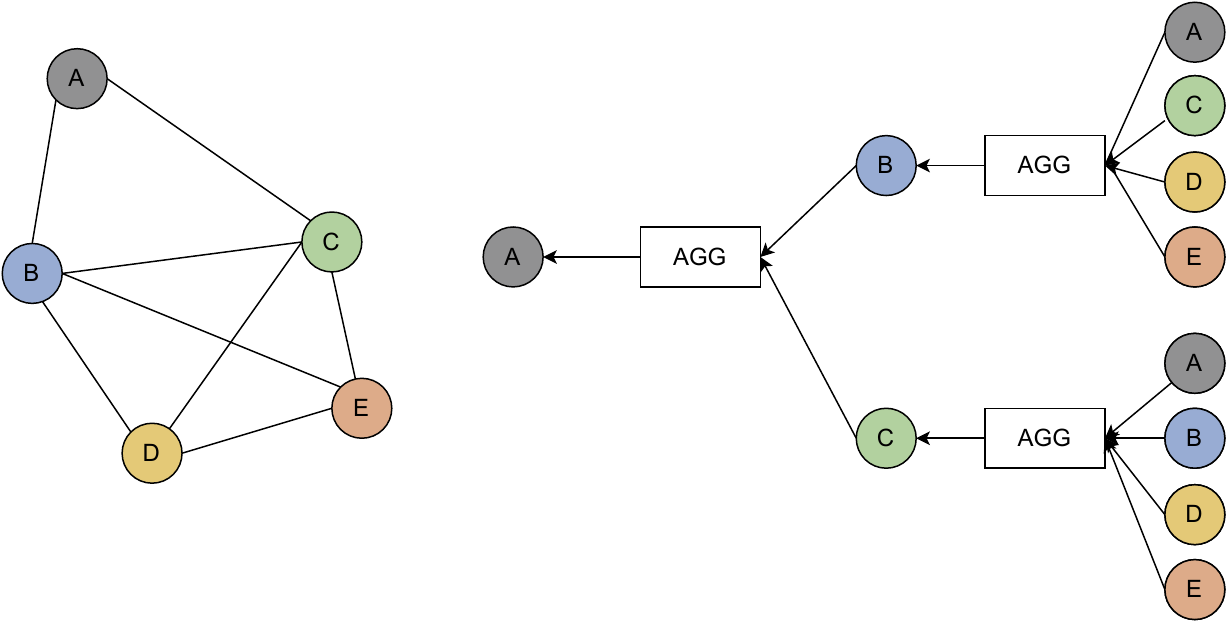}
    \caption{Two-hop graph convolutional networks with the full-neighbourhood sampling technique.} 
    \label{fig:graphsage_overview}
\end{figure}

This aggregation process is illustrated in Figure \ref{fig:graphsage_overview} (right); the k-hop neighbourhood node features of each graph node are aggregated based on the aggregation function.

The aggregated representation from the sampled neighbours is then concatenated with the node representation of itself $\mathbf{h}_v^{k-1}$. 
After that, the model weights $\mathbf{W}^k$ can be applied and the result is passed to a non-linear activation function $\sigma$ (e.g., ReLU) for obtaining the final node embeddings $\mathbf{h}_{v}^{k}$, as shown in Equation \ref{eq:general_graphsage} \cite{Hamilton2017}.

\begin{equation}
 \small
 \label{eq:general_graphsage}
 \mathbf{h}_{v}^{k} = \sigma\left(\mathbf{W}^{k}\cdot\operatorname{CONCAT}\left(\mathbf{h}_{v}^{k-1}, \mathbf{h}_{\mathcal{N}(v)}^{k}\right)\right) 
\end{equation}

Thus, the final node representation of node $v$ can be represented as $\mathbf{z}_v$, which depicts the node embeddings at the final layer $K$, as shown in Equation \ref{eq:final_embedding}.
For the purpose of node classification, the node embeddings (node representation) $\mathbf{z}_v$ can be fed to a sigmoidal neuron or softmax layer for classification tasks.

 \begin{equation}
 \label{eq:final_embedding}
     \mathbf{z}_v = \mathbf{h}_v^K,\ \ \ \ \ \forall vs. \in \mathcal{V}
 \end{equation}
%  where $K$ is the k-hop neighbour or number of graph convolutional layers. 
%  \vspace{0.5cm}
 
%  \item \textbf{Back Propagation}

\subsection{E-GraphSAGE}\label{Edge-Based GraphSAGE}
GNNs have been successfully applied in various application domains. However, these approaches mainly focus on node features for message propagation and are currently unable to consider edge features for edge classification. Therefore, the \mbox{E-GraphSAGE} algorithm was proposed to capture $k$-hop edge features for generating edge embeddings for the purpose of network intrusion detection. This provides the basis for computing the corresponding edge embeddings and enabling edge classification; that is, the classification of computer network flows as benign or attack flows. 

The goal of a NIDS is to detect malicious network flows. The problem can therefore be formulated as an edge classification problem in graph representation learning.
The original GraphSAGE algorithm only considers node features for message information prorogation rather than edge features~\cite{Hamilton2017}. However, most network attack datasets only consist of network flow features. Therefore, to capture edge information, sampling and aggregating at a $k$-hop depth needs to occur on edge features of the graph. Additionally, the final output should be edge embeddings for malicious network flow detection, rather than node embeddings. As a result, the E-GraphSAGE algorithm was proposed with these modifications, shown in Algorithm~\ref{alg:E-GraphSAGE}.

%AAAAAAAAAAAAAAAAAAAAAAAAAAAAAAAAAAAAAAAAAAAA
\IncMargin{1.5em}
\begin{algorithm}[!t]\footnotesize
\SetKwData{This}{this}\SetKwData{Up}{up}
  \SetKwFunction{Union}{Union}\SetKwFunction{FindCompress}{FindCompress}
  \SetKwInOut{Input}{input}\SetKwInOut{Output}{output}

\Indp\Indpp
  \Input{
  Graph $\mathcal{G}(\mathcal{V}, \mathcal{E})$;\\ 
  input edge features $\left\{\mathbf{e}_{uv}, \forall {uv} \in \mathcal{E}\right\}$;\\
  input node features $\mathbf{x}_{v} = \{1, \ldots, 1\}$;\\ 
  depth $K$;\\ 
  weight matrices $\mathbf{W}^{k}, \forall k \in\{1, \ldots, K\}$;\\ non-linearity $\sigma$;\\ 
  differentiable aggregator functions ${AGG}_{k}$ ;
  \vspace{0.05cm}
  }
  
  \Output{
  Edge embeddings $\mathbf {z}_{uv}, \forall  {uv} \in \mathcal{E}$
    }
  \BlankLine
    $\mathbf{h}_{v}^{0} \leftarrow \mathbf{x}_{v}, \forall vs. \in \mathcal{V}$

  \For{$k\leftarrow 1\ \KwTo\ K$}{
    \For{$v \in \mathcal{V}$}{\label{forins}
    ${\mathbf{h}_{\mathcal{N}(v)}^{k} \leftarrow   \text { AGG}_{k}\left(\left\{\mathbf{h}_{u}^{k-1} \| \mathbf{e}_{uv}^{k-1}, \forall {u} \in \mathcal{N}(v),  uv \in \mathcal{E}\right\}\right) }$
     
     ${\mathbf{h}_{v}^{k}  \leftarrow \sigma\left(\mathbf{W}^{k}\cdot\operatorname{CONCAT}\left(\mathbf{h}_{v}^{k-1}, \mathbf{h}_{\mathcal{N}(v)}^{k}\right)\right) }$
     }
}
    $\mathbf{z}_v = \mathbf{h}_v^K$

  \For{$uv \in \mathcal{E}$}{
    ${\scriptstyle \mathbf{z}_{uv}^{K}  \leftarrow {CONCAT}\left(\mathbf{z}_{u}^{K}, \mathbf{z}_{v}^{K}\right)}$
}
\Return   $\mathbf{z}_{uv}^{K}$,   $\mathbf{z}_{v}$

\caption{E-GraphSAGE edge embedding \cite{lo2021graphsage}.}
\label{alg:E-GraphSAGE}
\Indp\Indpp
\end{algorithm}\DecMargin{1em}

%AAAAAAAAAAAAAAAAAAAAAAAAAAAAAAAAAAAAAAAAAAAA

The main differences when comparing the original GraphSAGE algorithm \cite{Hamilton2017} with E-GraphSAGE are the algorithm inputs, the message propagation function, and the algorithm outputs. 

The algorithm's input should include edge features $\left\{\mathbf{e}_{uv}, \forall {uv} \in \mathcal{E}\right\}$ for edge feature message propagation, which is missing from the original GraphSAGE. 
Since most network intrusion detection datasets only consist of edge information, the algorithm only uses edge features (network flows) for network intrusion detection. To initialise the node representations, node features are set to $\mathbf{x}_{v} = \{1, \ldots, 1\}$ for each node. The dimensions of these constant vectors also match the dimensions of the edge feature set, as shown in line 1 of the algorithm. 

In line 4, to perform neighbour information aggregation based on edge features rather than node features, a new neighbourhood aggregator function was proposed to create the \textit{aggregated edge features of the sampled neighbourhood edges and neighbour information} at the $k$-th layer, as shown in Equation~\ref{eq:general_edge_aggregator}.

\begin{equation}
\small
\label{eq:general_edge_aggregator}
\mathbf{h}_{\mathcal{N}(v)}^{k} \leftarrow \operatorname{AGG}_{k}\left(\left\{\mathbf{h}_{u}^{k-1} \| \mathbf{e}_{u v}^{k-1}, \forall u \in \mathcal{N}(v), u vs. \in \mathcal{E}\right\}\right)
\end{equation}

Where $\mathbf{e}_{uv}^{k-1}$ are represented as edge features $uv$ from $\mathcal{N}(v)$, the sampled neighbourhood of node $v$, at layer \mbox{$k$-$1$}. 
The aggregated edge features can be concatenated with neighbouring node embeddings $\mathbf{h}_{u}^{k-1}$ for edge feature message propagation.  
In line 5, the node embedding for node $v$ at layer $k$, based on $k$-hop edge features, is computed, and the neighbouring node embedding consists of the neighbouring edge features. Thus, the $k$-hop edge features and topological patterns in the graph are collected and aggregated, and can be combined with the node representation of the current node $\mathbf{h}_{v}^{k-1}$. 

The final node representations at depth $K$, $\mathbf{z}_v = \mathbf{h}_v^K$, are computed in line 6.  The final edge embeddings $\mathbf{z}_{uv}^{K}$ for each edge, $uv$, are computed as the concatenation of a pair of node embeddings, $u$ and $v$, as shown in Equation~\ref{eq:final_edge_embedding}.

\begin{equation}
\small
\label{eq:final_edge_embedding}
\mathbf{z}_{uv}^{K} = {CONCAT}\left(\mathbf{z}_{u}^{K}, \mathbf{z}_{v}^{K}\right), uv \in \mathcal{E}
\end{equation}
This represents the final output of the forward propagation stage in E-GraphSAGE.

\subsection{Deep Graph Infomax}
    \begin{figure}[h]
        \centering
        \includegraphics[width=0.8\columnwidth]{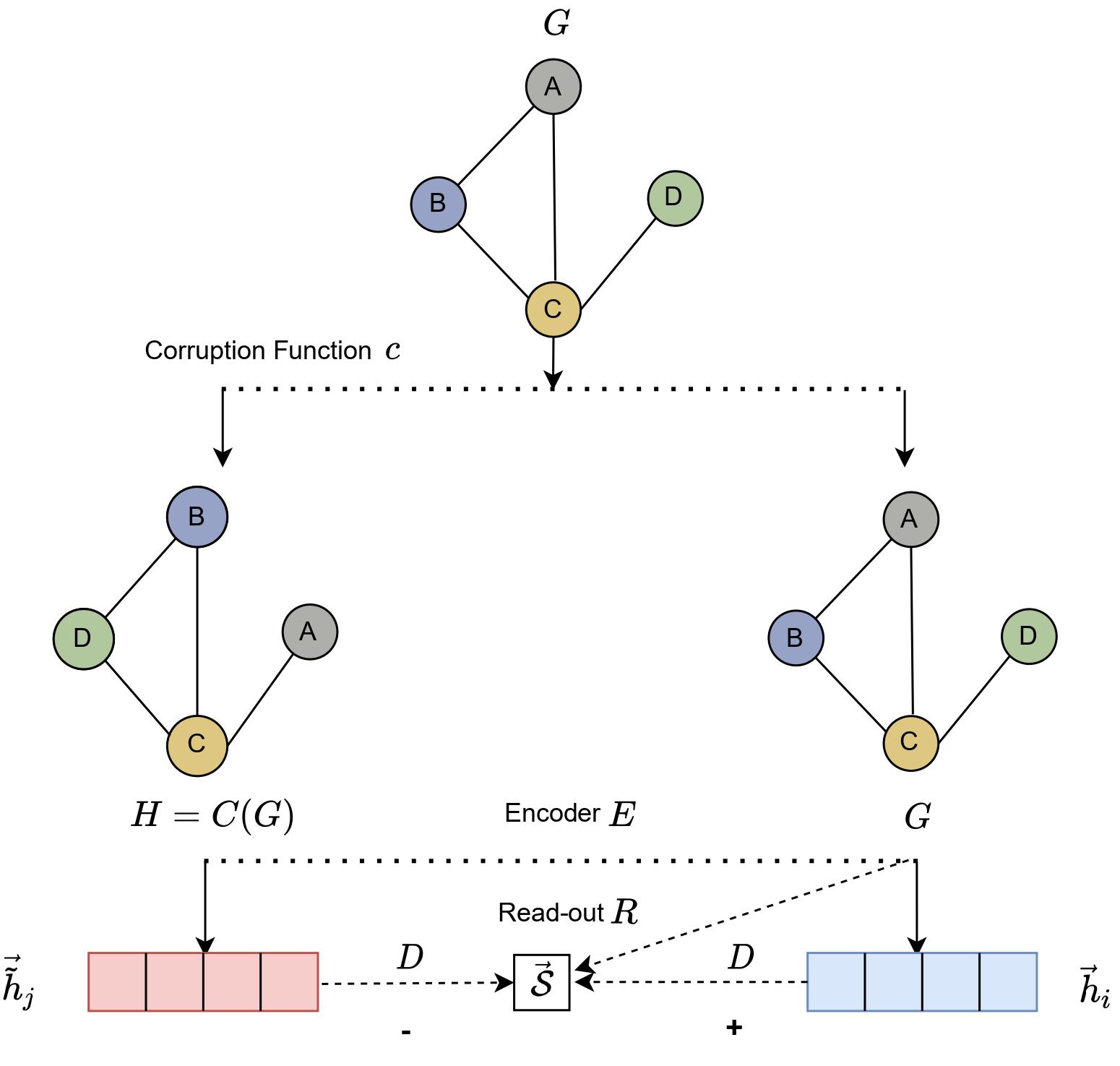}
        \caption{High-level overview of the DGI model \cite{lo2022inspection}.}
        \label{fig:dgi-overview}
    \end{figure}
    
Deep graph infomax (DGI) \cite{velivckovic2018deep} is a self-supervised graph-based learning approach that aims to train an encoder to learn node representations by maximising mutual information between patch representations and the global representation. Figure \ref{fig:dgi-overview} shows the overall process of DGI to perform mutual information maximisation. Here, $G$ represents the input graph used and $E$ represents the encoder. To train this encoder, a corrupted graph, $H$, is first created via a corruption function, ${C}$. 
 
As DGI focuses on node representations within the graph, the corruption function (i.e., random permutation of the input graph node features which add or remove edges from the adjacency matrix $A$) needs to be defined to generate corrupted graphs. These graphs are then passed into the encoder $E$, which can be any existing GNN, e.g., GraphSAGE. The encoder then outputs node embeddings of the input graph sample, $G$, and the negative graph sample, $H$.

With the input graph embeddings $G$, a global graph summary ${\overrightarrow{s}}$ is produced using a readout function ${R}$. ${\overrightarrow{s}}$ represents the summary of the input graph calculated by averaging all the given node embeddings and passing them through a sigmoid function. Finally, both the input graph embeddings ${\overrightarrow{h}_{i}}$ and negative sample embeddings ${\overrightarrow{\tilde h}_{j}}$ are evaluated through the use of a discriminator ${D}$. The discriminator uses the input graph and negative graph embeddings to score them against the global graph summary using a simple bilinear scoring method. By scoring these embeddings using the global graph summary, DGI achieves its goal in maximising the mutual local-global information and can update the parameters in ${D}$, ${R}$, and $E$, using gradient descent to continue training the encoder.

\section{Anomal-E}\label{anoml-E}
We present Anomal-E to address problems which existing graph-based NIDSs encounter, through their common reliance on supervised learning and node embeddings. As we aim to leverage edge-level information and learn edge embeddings in a self-supervised manner, Anomal-E utilises key features from two existing deep graph-based learning methodologies, E-GraphSAGE and DGI.

\begin{figure*}[!th]
    \raggedleft 
        \includegraphics[width=\textwidth]{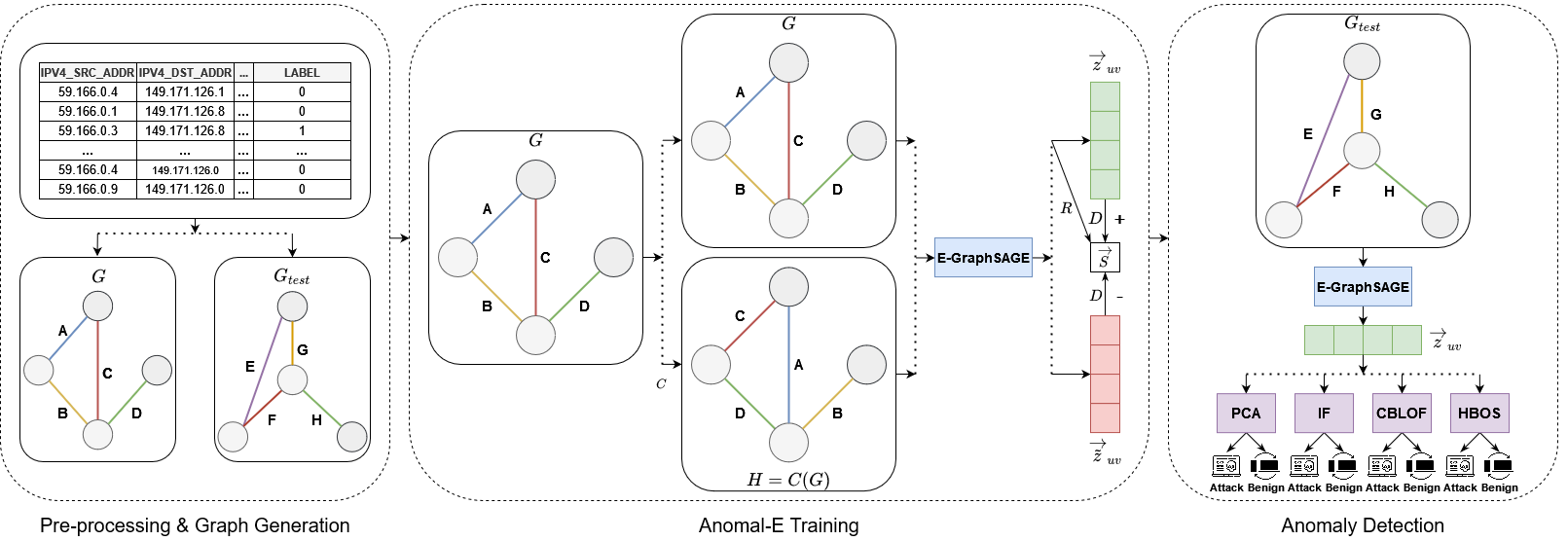}
        %\label{rfidtest_yaxis}
    \caption{Proposed Anomal-E NIDS pipeline. A given flow-based dataset is pre-processed into training and testing graphs; endpoints represent nodes and flows are the edges between them (left). The training graph is used to then train the Anomal-E model in which positive and negative embeddings are generated and used to tune and optimise the E-GraphSAGE model (middle). Finally, test embeddings are generated from the tuned E-GraphSAGE model to be used in downstream anomaly detection algorithms (right).}
    \label{fig:proposed_method}
\end{figure*}

    %AAAAAAAAAAAAAAAAAAAAAAAAAAAAAAAAAAAAAAAAAAAA
\IncMargin{1.5em}
\begin{algorithm}[!ht]\footnotesize
\SetKwData{This}{this}\SetKwData{Up}{up}
  \SetKwFunction{Union}{Union}\SetKwFunction{FindCompress}{FindCompress}
  \SetKwInOut{Input}{input}\SetKwInOut{Output}{output}
    \SetKwInOut{Input}{input}\SetKwInOut{Output}{output}

\Indp\Indpp
  \Input{
  Graph $\mathcal{G}(\mathcal{V}, \mathcal{E}, \mathcal{A},\mathcal{X})$;\\ 
  Input edge features $\left\{\mathbf{e}_{uv}, \forall {uv} \in \mathcal{E}\right\}$;\\
  Input node features $\mathbf{x}_{v} = \{1, \ldots, 1\}$;\\ 
  Depth $K$;\\ 
  Number of training epochs  $T$;\\ 
  Corruption function  $C$;\\ 

  \vspace{0.05cm}
  }
  
  \Output{
  Optimized E-GraphSAGE encoder $g$,  Optimized PCA $h_{-} PCA $ ,  Optimized Isolation Forest $h_{-} IF $, Optimized CBLOF $h_{-} CBLOF $,  Optimized HBOS $h_{-} HBOS $
    }
  \BlankLine
    Initialize the parameters $\theta$ and $\omega$ for the encoder $g$ and the discriminator $D$;

  \For{$epoch\leftarrow 1\ \KwTo\ T$}{

    $\mathbf{z}_{uv}^{K}, \mathbf{z}_{v}^{K} =g(G, \theta)$ \\
    $\widetilde{z}_{uv}^{K}, \textunderscore=g(C(G), \theta) $\\
    $\bar{s}= \sigma\left(\frac{1}{n} \sum_{i=1}^{
    n} z_{v}^{(K)}\right)$
     
    $\mathcal{D}\left( \mathbf{z}_{uv}^{K} , \bar{s}\right)=\sigma\left(\mathbf{z}_{uv}^{K^T} \mathbf{w} \bar{s}\right)$\\

    $\mathcal{D}\left( \widetilde{z}_{uv}^{K} , \bar{s}\right)=\sigma\left(\widetilde{z}_{uv}^{T} \mathbf{w} \bar{s}\right)$ \\

$\mathcal{L}_{DGI}=-\frac{1}{2 n} \sum_{i=1}^{n}\left(\mathbb{E}_{G} \log \mathcal{D}\left(\boldsymbol{z}_{uv}^{(K)}, \bar{s}\right)+\mathbb{E}_{\tilde{G}} \log \left(1-\mathcal{D}\left(\tilde{\boldsymbol{z}}_{uv}^{(K)}, \bar{s}\right)\right)\right)$

    $\theta, \omega \leftarrow \operatorname{Adam}\left(\mathcal{L}_{DGI}\right)$
     }
    $\mathbf{z}_{uv}^{K}, \textunderscore =g(G, \theta)$ \\
    $h_{-} IF \leftarrow \mathrm{IF}(z_{uv})$\\
    $h_{-} PCA \leftarrow \mathrm{PCA}(z_{uv})$\\
    $h_{-} CBLOF \leftarrow \mathrm{CBLOF}(z_{uv})$\\
    $h_{-} HBOS \leftarrow \mathrm{HBOS}(z_{uv})$\\

\Return   $h_{-} IF, h_{-} PCA, h_{-} CBLOF, h_{-} HBOS, g$

\caption{Pseudocode of the Anomal-E algorithm.}
\label{alg:Anoml-E}
\Indp\Indpp
\end{algorithm}\DecMargin{1em}

%AAAAAAAAAAAAAAAAAAAAAAAAAAAAAAAAAAAAAAAAAAAA    

        Figure \ref{fig:proposed_method} presents an overview of the anomaly detection process performed using Anomal-E. Firstly, raw network flows are pre-processed to generate representative training and testing graphs. The training graph is then fed into the \mbox{Anomal-E} training process to train the E-graphSAGE encoder. By using \mbox{E-GraphSAGE}, Anomal-E aims to learn edge representations that maximise local-global mutual information in a self-supevised context. Similarly to the DGI methodology, a negative graph representation is first generated using a corruption function. However, rather than generating a random permutation of the input graph's node features, edge features are randomly permutated instead. Both of these graphs are then passed through the E-GraphSAGE encoder. E-GraphSAGE then outputs the edge embeddings of the input graph and negative graph. As with the original DGI methodology, these embeddings are used to generate a global graph summary so that the input and negative embeddings can be scored against using a discriminator. Now, parameter adjustment occurs using these scores, and the process is repeated to continue training the E-graphSAGE encoder. Once training is complete, the output training graph embeddings are used to train four different existing anomaly detection algorithms. Detection accuracy is then determined through inputting the testing graph through these algorithms. These steps are explained in more detail in the following sections.

           \subsection{Algorithm Description}
           Anomal-E is represented in Algorithm \ref{alg:Anoml-E}. For training the proposed model, the input includes the network graph $\mathcal{G}$, with network flow features to extract true edge embeddings and corrupted edge embeddings. For this, we need to define the corruption function $\mathcal{C}$  to generate the corrupted network graphs $\mathcal{C(G)}$ for E-GraphSAGE to extract the corrupted edge embeddings. In this paper, we randomly shuffled the edge features among the edges in the real network graph $\mathcal{G}$, to generate the corrupted network graphs. This is done by shuffling the edge feature matrix in rows $\mathbf{X}$. Overall, instead of adding or removing edges from the adjacency matrix such that $(\widetilde{\mathbf{A}} \neq \mathbf{A})$, we use corruption function $\mathcal{C}$, which shuffles the edge features such that $(\tilde{\mathbf{X}} \neq \mathbf{X})$, but retains the adjacency matrix $(\widetilde{\mathbf{A}} =\mathbf{A})$.  
           
           For each training epoch, lines 3 to 4, we use E-GraphSAGE to extract true edge embeddings, true global graph summaries, and corrupted edge embeddings; rather than having a 2-layer E-GraphSAGE model, a 1-layer model is used. This is because the base DGI model benefits from wider rather than deeper models.

           In lines $5$ to $6$, for the discriminator $\mathcal{D}$, we introduced the non-linear logistic sigmoid function, which compares the edge embedding vector $\vec{z}_{uv}$ against a real whole graph embedding $\overline{s}$ to calculate the score of $\mathcal(\vec{z}_{uv},\overline{s})$, being positive or negative. 
            
            \begin{equation}
            \mathcal{D}\left(\mathbf{z}_{uv}^{K} , \bar{s}\right)=\sigma\left(\mathbf{z}_{uv}^{K^T}  \mathbf{w} \bar{s}\right)
            \end{equation}
            
            \begin{equation}
            \mathcal{D}\left(\mathbf{\widetilde{z}}_{uv}^{K}, \bar{s}\right)=\sigma\left(\mathbf{\widetilde{z}}_{uv}^{K^T} \mathbf{w} \bar{s}\right)
            \end{equation}

            We then apply a binary cross-entropy loss objective function to perform gradient descent, as shown in line $8$. To perform gradient descent, we maximise the score if the edge embedding is a true edge embedding $\vec{z}_{uv}$ and minimise the score if it is a corrupted edge embedding $\vec{z}_{uv}$ compared to the global graph summary generated by the read-out function $\mathcal{R}$. As a result, we can maximise the mutual information between patch representations (i.e., edge representations) and the whole real graph summary. After the training process, the trained encoder can be used to generate new edge embeddings for downstream purposes, i.e., anomaly detection. Finally, in line $11$ to $14$ we applied various anomaly detection algorithms, including PCA, IF, CBLOF, and HBOS, for unsupervised anomaly detection.

           \subsection{Pre-processing and Graph Generation}
\begin{figure}[!th]
                    \centering
                    \includegraphics[width=0.98\columnwidth]{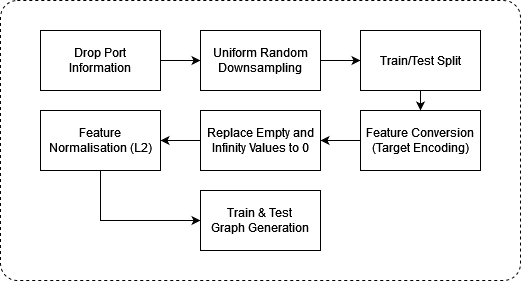}
                    %\label{rfidtest_yaxis}
                \caption{Pre-processing and graph generation method.}
                \label{fig:preprocess}
            \end{figure}

            Before training, data are first converted from its raw Netflow format \cite{rfc3954} into its equivalent graph-based format. NetFlow is an IP flow gathering tool leveraging network elements, such as routers and switches, to gather encountered IP flows and export them to external devices. These IP flows can be defined as unidirectional sequences of packets that are encountered on network devices and contain a variety of important network information, including IP addresses, port numbers, packet and byte counts, and other useful packet statistics. Netflow data allow for a seamless integration of flow records into graph format, as nodes can be represented through IP addresses and/or ports, and flow information can naturally represent edges. Figure \ref{fig:preprocess} shows an overview of the process undertaken to pre-process Netflow records from each dataset into training and testing graphs. For experiments in this study, source and destination port information is firstly removed from each flow record. The data are then downsampled to 10\% of their original size, similarly to what was seen in \cite{singla2020preparing}. On downsampled flows, samples are separated into training and testing sets. Target encoding on categorical features in both the training and testing set then occurs. To perform this in an unsupervised manner, the training set is used to train a target encoder on categorical data. Encoding is then performed again using the fully trained encoder on both the training and testing set. Any empty or infinite values arising from this process is replaced with a value of 0. Before finally generating graphs, training and testing sets are normalised using an L2 normalisation approach. Similarly to the approach taken in target encoding, the normaliser is trained using the training set to ensure an unsupervised process is achieved. Finally, the training and testing sets are converted into bidirectional graph representations. Processed flow information is then applied as edge features. Additionally, following the E-GraphSAGE methodology, each graph's node features are set to a constant vector containing \textit{ones} with the same dimensions as those of the edge features; i.e., if 49 edge features are used, node features will be vectors of 49 ones. 

        \subsection{Training}
            For each dataset and experiment, training occurred using the generated training graph, as previously discussed. Before training commenced in the Anomal-E model, the encoder was set to E-GraphSAGE. The encoder uses a mean aggregation function and uses similar hyperparameters to the approach taken by Lo et al. \cite{lo2021graphsage} in their paper; however, rather than having a 2-layer E-GraphSAGE model, a 1-layer model is used. Hyperparameter selection can be seen in Table \ref{anomal-e_parameters}. To summarise, the encoder uses a hidden layer size of 128 units, which also matches the final output layer size. However, as the encoder generates edge embeddings, the size is doubled to 256. ReLU is used for the activation function, and no dropout rate is included. As for the generation of the global graph summary, we follow the same approach used by the authors of the DGI paper. However, instead of averaging node embeddings and passing them through a sigmoid function, edge embeddings are used. A binary cross entropy (BCE) function is used to calculate loss, and gradient descent is used for backpropagation with the Adam optimiser using a learning rate of 0.001.
            \begin{table}[!ht]\small
            \renewcommand{\arraystretch}{1.3}
            \caption{Hyperparameter values used in Anomal-E.}
            \label{anomal-e_parameters}
            \centering
            \scalebox{0.8}{ 
            \begin{tabular}{*2l }
                \toprule
                 
                Hyperparameter & Values  \\
                \toprule
                
                No. Layers & 1 \\
                No. Hidden & $256$\\
                Learning Rate & $1e^{-3}$\\
                Activation Func. & ReLU\\
                Loss Func. & BCE\\
                Optimiser & Adam\\
            \end{tabular}
            }
        \end{table}
        \subsection{Anomaly Detection}\label{anom-detect}
            Upon completion of training, the tuned E-GraphSAGE model was used to generate edge embeddings from both the training and testing graphs. To perform anomaly detection, the four algorithms PCA, IF, CBLOF, and HBOS were used. To train each algorithm in an unsupervised manner, edge embeddings generated from the training graph were used as inputs. Contaminated and non-contaminated training samples were also considered in the experiments, i.e., training with attack samples omitted and training with attack samples included. Grid searches were also performed in each experiment for each detection algorithm to ensure optimal parameter tuning. All algorithms were grid searched on the contamination parameter along with number of components for PCA, number of estimators for IF, number of clusters for CBLOF, and number of bins for HBOS. % Please ensure intended meaning is retained 
          Table \ref{grid_search} displays the grid search parameters in greater detail. Once each algorithm was trained, the test edge embeddings were then used to evaluate the success of the model in an unsupervised fashion, as seen in the "Anomaly Detection" phase in Figure \ref{fig:proposed_method}. 
            
            \begin{table}[!ht]\small
            \renewcommand{\arraystretch}{1.3}
            \caption{Hyperparameter values used in grid search for anomaly detection algorithms.}
            \label{grid_search}
            \centering
            \scalebox{0.8}{ 
            \begin{tabular}{*3l }
                \toprule
                 
                Algorithm & Hyperparameter & Values  \\
                \toprule
                
                PCA & No. components & $[5, 10, 15, 20, 25, 30]$\\
                IF & No. estimators & $[20, 50, 100, 150]$\\
                CBLOF & No. clusters & $[2, 3, 5, 7, 9, 10]$\\
                HBOS & No. bins & $[5, 10, 15, 20, 25, 30]$\\
                \toprule
                All & contamination & $[0.001, 0.01, 0.04, 0.05, 0.1, 0.2]$\\
            \end{tabular}
            }
        \end{table}
            
            It is important to note that the PCA-based algorithm  \cite{shyu2003novel} used at this stage for anomaly detection was an extension of the regular PCA algorithm, adapted for anomaly detection rather than dimensionality reduction. This method first performs a PCA reduction on a set of "normal" samples to extract a correlation matrix. This "normal" correlation matrix then acts as a criterion for the outlier detection stage of the algorithm; i.e., if a sample deviates to a certain degree from this normal, it is considered an outlier. In this paper, we refer to this extension of the PCA algorithm simply as PCA.
            
            In IF anomaly detection~\cite{liu2008isolation}, tree structures are leveraged to isolate and identify samples that represent anomalies. If a sample is isolated close to the root of the tree, then it is identified as an anomaly, and samples deeper in the tree are determined as "normal." This process is completed with an ensemble of trees to create an isolation forest. This commonly used algorithm is found to identify anomalies effectively and accurately at an efficient rate.
            
            The CBLOF~\cite{he2003discovering} algorithm approaches anomaly detection as a clustering-based problem. Samples are clustered and given an outlier factor that represents both the size of the cluster that the sample belongs to, and the distance between itself and the nearest cluster. This factor defines the degree of variation the sample has, which is used to determine if it is an anomaly or not.
            
            HBOS~\cite{goldstein2012histogram}, on the other hand, takes a histogram-based approach to anomaly detection, rather than the cluster-based approach taken for CBLOF. In this method, a univariate histogram is constructed for each feature; each histogram contains a set of bins. The frequency of slotted samples in each bin for each histogram enables for bin densities to be derived. The density value calculated for each histogram is then used to determine the HBOS score to dictate whether a sample is an outlier.

\section{Experiments and Results}\label{results}
    % Approach also discussed?
    For the performance evaluation of Anomal-E, multiple testing approaches were used. As labelled datasets were used, both contaminated and non-contaminated datasets were investigated and evaluated against multiple standard anomaly detection algorithms. Datasets were randomly and uniformly downsampled to $10\%$ due to the large size of their original representations. In training, $70\%$ of the data were used, and the remaining $30\%$ of samples were used for testing and performance evaluation. 
    
    \subsection{Datasets}
        Two benchmark NIDS datasets were used in our experiments: NF-UNSW-NB15-v2 and NF-CSE-CIC-IDS2018-v2. These are both updated versions of existing NIDS datasets that have been standardised into a NetFlow format \cite{rfc3954} by Sarhan et al. \cite{sarhan2022towards}. Further details on each dataset are as follows:
        \begin{itemize}
            \item NF-UNSW-NB15-v2 \\NetFlow formatted version of UNSW-NB15 \cite{7348942} with 43 standardised features. Out of the 2,390,275 flows, 2,295,222 (96.02\%) samples are benign flows, and the remaining 95,053 (3.96\%) samples are attack flows. Nine attack types are distributed across the 3.98\% of flows representing attacks.
            \item NF-CSE-CIC-IDS2018-v2 \\Adapted from the CSE-CIC-IDS2018 \cite{icissp18}. This dataset contains 18,893,708 flows distributed between 16,635,567 (88.05\%) benign samples and 2,258,141 (11.95\%) attack samples. Within the 11.95\% of attack samples are 6 different attack methods.
        \end{itemize}
    
    Non-contamination experiments omitted all attack samples during training, whereas 4\% of attack samples were included for both datasets in contamination experiments. This is due to NF-UNSW-NB15-v2 having a natural attack contamination level of approximately 4\%. Therefore, to ensure experimental equality, NF-CSE-CIC-IDS2018-v2 was contaminated to the same degree. 
    
    \subsection{Evaluation metrics}\ \label{eval}
    To clearly distinguish the differences and benefits of using Anomal-E embeddings in anomaly detection, our results compare macro F1-scores between using raw features and Anomal-E embeddings as anomaly detection algorithm inputs. Macro F1-scores (Macro F1) are used as a key performance indicator for our experiments, as this metric combats the problem of potential imbalance within datasets. Additionally, accuracy (Acc) and detection rate (DR) are used for further performance evaluation.
    
    \subsection{NF-UNSW-NB15-v2 Results}
\begin{figure}[!ht]
            \centering
            \includegraphics[width=\columnwidth]{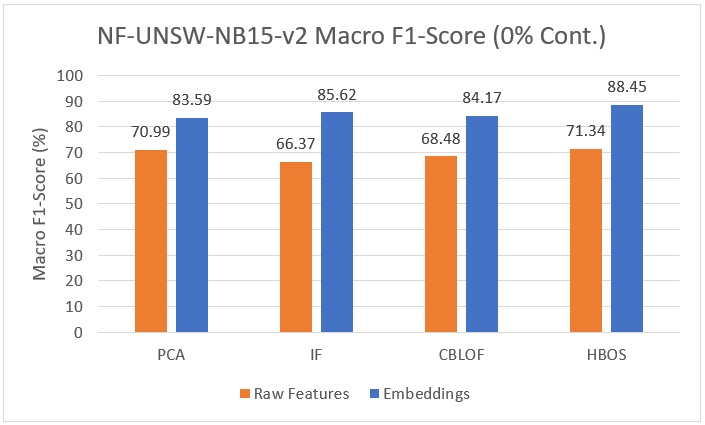}
            \caption{NF-UNSW-NB15-v2 macro F1-score (0\% contamination) comparison.}
            \label{graph_unsw_0_f1}
        \end{figure}

        Table \ref{unsw_table_no_contamination} presents results of the experiments on NF-UNSW-NB15-v2 without dataset contamination. The results clearly display the benefit of using Anomal-E embeddings according each performance metric and for each anomaly detection algorithm. In particular, the macro F1-scores seen in Figure \ref{graph_unsw_0_f1} clearly display the significant improvement of using Anomal-E embeddings over raw features, for all anomaly detection methods. For example, when using raw embeddings, a score of only 71.34\% was achieved in the best case. In the best case when using Anomal-E, a score of 88.45\% was achieved. This establishes the benefit of using Anomal-E for network anomaly detection.
        
\begin{figure}[!ht]
            \centering
            \includegraphics[width=\columnwidth]{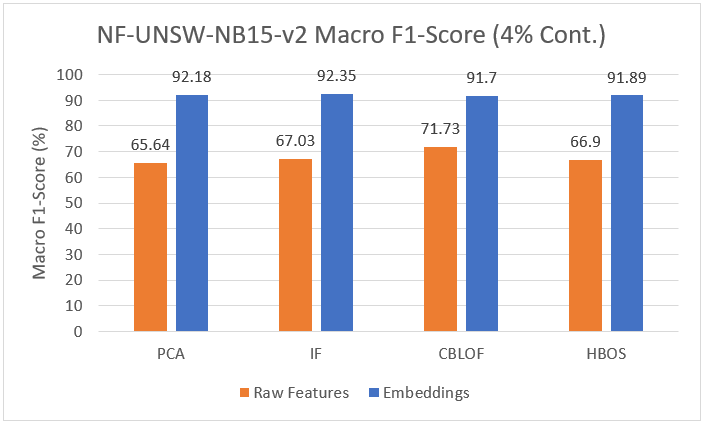}
            \caption{NF-UNSW-NB15-v2 Macro F1 (4\% contamination) comparison.}
            \label{graph_unsw_4_f1}
        \end{figure}

         \begin{table}[!ht]\small
            \renewcommand{\arraystretch}{1.3}
            \caption{NF-UNSW-NB15-v2 results (0\% contamination).}
            \label{unsw_table_no_contamination}
            \centering
            \scalebox{0.8}{
            \begin{tabular}{c | *3c | *3c}
                \toprule
                \multicolumn{1}{c}{} &  \multicolumn{3}{c}{Raw Features} & \multicolumn{3}{c}{Embeddings} \\ 
                {} & Acc & Macro F1 & DR & Acc & Macro F1 & DR \\
                PCA & $94.45\%$ & $70.99\%$ & $56.87\%$ & $97.64\%$ & $83.59\%$ & $64.20\%$ \\
                IF & $89.64\%$ & $66.37\%$ & $81.17\%$ & $97.91\%$ & $85.62\%$ & $68.76\%$ \\
                CBLOF & $94.12\%$ & $68.48\%$ & $49.37\%$ & $97.70\%$ & $84.17\%$ & $65.97\%$ \\
                HBOS & $94.47\%$ & $71.34\%$ & $58.24\%$ & $98.18\%$ & $88.45\%$ & $80.36\%$ \\
            \end{tabular}
            }
        \end{table}
    Results for NF-UNSW-NB15-v2 with dataset contamination are presented in Table \ref{unsw_table_contamination}; macro F1-scores are visualised in Figure \ref{graph_unsw_4_f1}. A similar trend can be seen to those presented in the non-contamination version of experiments. The use of raw features causes macro F1-scores and detection rates to be significantly lower than when using Anomal-E embeddings. Again, embeddings achieved consistently higher results in all metrics for all detection algorithms. It can also be seen that for raw features that the macro F1-scores on average, for all detection algorithms, are lower than the equivalent experimental results using no contamination. In contrast, the use of Anomal-E embeddings achieved consistently high results for all four algorithms, and even improved on embeddings results seen in non-contamination experiments. Again, this demonstrates the benefits of Anomal-E and leveraging graph structures in anomaly detection compared to using only raw features.
        \begin{table}[!ht]
            \renewcommand{\arraystretch}{1.3}
            \caption{NF-UNSW-NB15-v2 results (4\% contamination).}
            \label{unsw_table_contamination}
            \centering
             \scalebox{0.72}{
            \begin{tabular}{c | *3c | *3c}
                \toprule
                \multicolumn{1}{c}{} &  \multicolumn{3}{c}{Raw Features} & \multicolumn{3}{c}{Embeddings} \\ 
                {} & Acc & Macro F1 & DR & Acc & Macro F1 & DR \\
                PCA & $91.11\%$ & $65.64\%$ & $63.01\%$ & $98.63\%$ & $92.18\%$ & $97.86\%$ \\
                IF & $91.46\%$ & $67.03\%$ & $67.60\%$ & $98.66\%$ & $92.35\%$ & $98.77\%$ \\
                CBLOF & $95.28\%$ & $71.73\%$ & $50.42\%$ & $98.57\%$ & $91.70\%$ & $95.72\%$ \\
                HBOS & $91.59\%$ & $66.90\%$ & $65.67\%$ & $98.62\%$ & $91.89\%$ & $94.92\%$ \\
            \end{tabular}
            }
        \end{table}
        
        % \begin{figure}[!ht]
        %     \centering
        %     \includegraphics[width=\columnwidth]{figures/UNSW-4-all.PNG}
        %     \caption{NF-UNSW-NB15-v2 Results (4\% contamination) Comparison}
        %     \label{graph_unsw_4_all}
        % \end{figure}
        
   \subsection{NF-CSE-CIC-IDS2018-v2 Results}
\begin{figure}[!ht]
            \centering
            \includegraphics[width=\columnwidth]{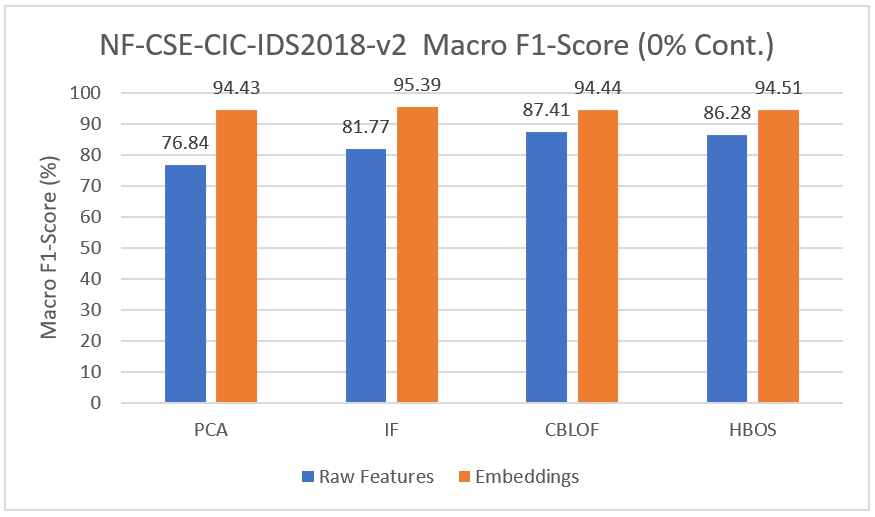}
            \caption{NF-CSE-CIC-IDS2018-v2 Macro F1 (0\% contamination) comparison.}
            \label{graph_cic_0_f1}
        \end{figure}

        Similarly to NF-UNSW-NB15-v2 results, Table \ref{cic_table_no_contamination} displays results for anomaly detection on NF-CSE-CIC-IDS2018-v2 without dataset contamination. For all anomaly detection algorithms, Anomal-E embeddings outperformed raw features for accuracy and macro F1-score, as seen in Figure \ref{graph_cic_0_f1}. The overall level and consistency of improvements when using Anomal-E embeddings over raw features demonstrate the benefit of Anomal-E's ability to leverage and generalise graph structure and edge features for network anomaly detection with no contamination. High detection rates and accuracy, in conjunction with Anomal-E's self-supervised training methodology, also display the potential this approach has for detecting network anomalies in different attack datasets and environments.
        
\begin{figure}[!ht]
            \centering
            \includegraphics[width=\columnwidth]{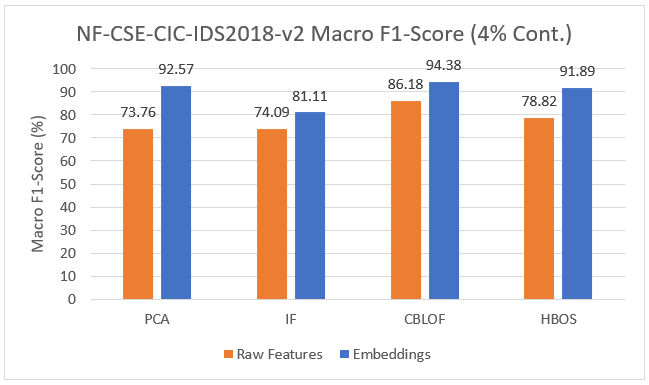}
            \caption{NF-CSE-CIC-IDS2018-v2 Macro F1 (4\% contamination) comparison.}
            \label{graph_cic_4_f1}
        \end{figure}

        \begin{table}[!ht]\small
            \renewcommand{\arraystretch}{1.3}
            \caption{NF-CSE-CIC-IDS2018-v2 results (0\% contamination).}
            \label{cic_table_no_contamination}
            \centering
            \scalebox{0.8}{
            \begin{tabular}{c | *3c | *3c}
                \toprule
                \multicolumn{1}{c}{} &  \multicolumn{3}{c}{Raw Features} & \multicolumn{3}{c}{Embeddings} \\ 
                {} & Acc & Macro F1 & DR & Acc & Macro F1 & DR \\
                PCA & $88.30\%$ & $76.84\%$ & $75.13\%$ & $97.82\%$ & $94.43\%$ & $82.67\%$ \\
                IF & $92.06\%$ & $81.77\%$ & $70.82\%$ & $98.18\%$ & $95.39\%$ & $85.38\%$ \\
                CBLOF & $94.42\%$ & $87.41\%$ & $82.78\%$ & $97.83\%$ & $94.44\%$ & $82.67\%$ \\
                HBOS & $94.01\%$ & $86.28\%$ & $79.28\%$ & $97.72\%$ & $94.51\%$ & $88.55\%$ \\

            \end{tabular}
            }
        \end{table}
        For results gathered using NF-CSE-CIC-IDS2018-v2 with dataset contamination, shown in Table \ref{cic_table_contamination}, we can observe that again, Anomal-E embeddings outperformed raw features. Results clearly show significant increases for all algorithms when using a contaminated NF-CSE-CIC-IDS2018-v2 dataset. In a similar fashion to observations made in contaminated experiments on NF-UNSW-NB15-v2, the scores across performance metrics and detection algorithms are worse for raw features (see Figure \ref{graph_cic_4_f1}) with the contaminated dataset compared to the non-contaminated one. When comparing contaminated and non-contaminated Anomal-E embeddings, we can observe an improvement when using the non-contaminated NF-CSE-CIC-IDS2018-v2 dataset. However, on average, contaminated Anomal-E embeddings still display improvements over non-contaminated raw feature results. This again emphasises the strong ability Anomal-E has for network anomaly detection. 
        \begin{table}[!ht]\small
            \renewcommand{\arraystretch}{1.3}
            \caption{NF-CSE-CIC-IDS2018-v2 results (4\% contamination).}
            \label{cic_table_contamination}
            \centering
            \scalebox{0.8}{
            \begin{tabular}{c | *3c | *3c}
                \toprule
                \multicolumn{1}{c}{} &  \multicolumn{3}{c}{Raw Features} & \multicolumn{3}{c}{Embeddings} \\ 
                {} & Acc & Macro F1 & DR & Acc & Macro F1 & DR \\
                PCA & $85.91\%$ & $73.76\%$ & $74.71\%$ & $97.11\%$ & $92.57\%$ & $79.16\%$ \\
                IF & $86.1\%$ & $74.09\%$ & $75.39\%$ & $89.79\%$ & $81.11\%$ & $91.84\%$ \\
                CBLOF & $94.61\%$ & $86.18\%$ & $69.16\%$ & $97.80\%$ & $94.38\%$ & $82.67\%$ \\
                HBOS & $88.81\%$ & $78.82\%$ & $84.22\%$ & $96.86\%$ & $91.89\%$ & $77.79\%$ \\
            \end{tabular}
            }
        \end{table}

   \subsection{Comparative results with baseline methods}
   
   In this section, we discuss the performance evaluation of our proposed framework and the comparison with several baseline methods on the NF-UNSW-NB-15-v2 and NF-CSE-CIC-IDS2018-v2 datasets. Specifically,  GraphSAGE~\cite{Hamilton2017} and DGI~\cite{velickovic2018deep} were used as the comparative baseline methods. However, for each baseline method, rather than leveraging edge features in the training process, node features were leveraged instead. This is because both methods, by default, can use node features to generate edge embeddings, and unlike Anomal-E, do not take edge features into account when training. Edge embeddings produced by each baseline were used in the same anomaly detection algorithms discussed in Section \ref{anom-detect} and Section \ref{results}. It is important to note that the combination of GraphSAGE or DGI with each anomaly detection algorithm was created for a baseline comparison. To the best of our knowledge, this paper was the first to combine these methods for network intrusion detection. Again, we evaluated performance through the same metrics used in Section~\ref{eval}. 
  Table \ref{state_art_no_contamination} and \ref{state_art_contamination} display these results for non-contaminated and contaminated experiments, respectively, on both datasets.

   In Table \ref{state_art_no_contamination}, Anomal-E achieved the highest detection accuracy across all performance metrics for experiments on NF-UNSW-NB15-v2. On average, for this dataset, anomaly detection using Anomal-E embeddings achieved a macro F1-score of 85.45\%. This is significantly higher than than the average macro F1-scores found when using the detection algorithms with DGI or GraphSAGE. This displays the benefit of leveraging edge features rather than node features for generating edge embeddings. Similarly, \mbox{Anomal-E} dominated the experiments on NF-CSE-CIC-IDS2018-v2. Overall, for non-contaminated experiments, Anomal-E still achieved dramatically better detection performance than the baseline methods. From the results, it is most important to note that our proposed method can be generalised to different network attack datasets. This is seen clearly through the Anomal-E-HBOS experiment. GraphSAGE-HBOS slightly outperformed Anomal-E-HBOS for experiments on the NF-CSE-CIC-IDS2018-v2 dataset. However, when observing average macro F1-score across datasets, GraphSAGE-HBOS dramatically reduced detection performance, reaching a score of 82.54\%. If we observe Anomal-E in the same capacity, we see that our Anomal-E-HBOS method improved on this with an average macro F1-score of 91.48\%. This trend was consistent across all other Anomal-E algorithm combinations.

   In Table \ref{state_art_contamination}, displaying experiments with contamination, there is a similar trend as that in Table \ref{state_art_no_contamination}. Again, anomaly detection using Anomal-E embeddings improved performance across all algorithms. For example, there was a considerable increase of 43.36\% in the macro F1-score when comparing Anomal-E-IF to DGI-IF on NF-UNSW-NB15-v2. As for experiments on NF-CSE-CIC-IDS2018-v2, we can observe some scenarios in which the GraphSAGE baseline method slightly surpassed Anomal-E's results. Specifically, this can be seen for GraphSAGE-CBLOF and GraphSAGE-HBOS. Although these baseline experiments just outperformed Anomal-E on this specific dataset, GraphSAGE failed to generalise to both datasets. When averaging macro F1-scores across both datasets, Anomal-E-CBLOF and Anomal-E-HBOS achieved scores of 93.04\% and 91.89\%, respectively, whereas GraphSAGE-CBLOF and GraphSAGE-HBOS achieved only 72.07\% and 74.69\%. This is similar to what we noticed in non-contamination experiments, in that the baseline only performed well on one dataset and performed significantly worse on the other. 

         \begin{table*}[!ht]
            \renewcommand{\arraystretch}{1.3}
            \caption{Performance evaluation for metrics on both datasets compared with the baseline algorithms (0\% contamination).}
            \label{state_art_no_contamination}
            \centering
            \scalebox{0.8}{
            \begin{tabular}{c | *3c | *3c | *1c}
                \toprule
                \multicolumn{1}{c}{} &  \multicolumn{3}{c}{NF-UNSW-NB15-v2} & \multicolumn{3}{c}{NF-CSE-CIC-IDS2018-v2} &
                \multicolumn{1}{c}{Average Across Datasets} \\ 
                {} & Acc & Macro F1 & DR & Acc & Macro F1 & DR & Macro F1 \\
                
                Anomal-E-PCA (Ours) & $\textbf{97.64\%}$ &  $\textbf{83.59\%}$ & $\textbf{99.63\%}$ & $\textbf{97.82\%}$ & $\textbf{94.43\%}$ & $\textbf{82.67\%}$ &  \textbf{89.01\%}  \\

                DGI-PCA & $96.02\%$ & $48.89\%$ & $0.00\%$ & $88.03\%$ & $46.82\%$ & $0.00\%$ & 47.86\% \\
                GraphSAGE-PCA & $95.95\%$ & $48.97\%$ & $0.00\%$ & $88.05\%$ & $46.82\%$ & $0.00\%$  & 47.90\%   \\

\hline
                 Anomal-E-IF (Ours) & $\textbf{97.91\%}$ & $\textbf{85.62\%}$ & $\textbf{68.76\%}$ & $\textbf{98.18\%}$ & $\textbf{95.39\%}$ & $\textbf{85.38\%}$ &  \textbf{90.51\%}  \\

                DGI-IF & $96.02\%$ & $48.99\%$ & $0.00\%$ & $88.03\%$ & $46.82\%$ & $0.00\%$ &  47.91\% \\
                GraphSAGE-IF & $96.02\%$ & $48.99\%$ & $0.00\%$ & $88.05\%$ & $46.82\%$ & $0.00\%$ &  47.91\% \\

\hline
                 Anomal-E-CBLOF (Ours) & $\textbf{97.70\%}$ & $\textbf{84.17\%}$ & $\textbf{65.97\%}$ & $\textbf{97.83\%}$ & $\textbf{94.44\%}$ & $82.67\%$  & \textbf{89.31\%} \\
                DGI-CBLOF & $96.02\%$ & $48.99\%$ & $0.00\%$ & $88.03\%$ & $46.82\%$ & $0.00\%$  & 47.91\% \\
                 
                GraphSAGE-CBLOF & $87.10\%$ & $49.53\%$ & $10.32\%$ & $90.74\%$ & $82.76\%$ & $\textbf{95.08\%}$  & 66.15\% \\
                
                \hline
                 Anomal-E-HBOS (Ours) & $\textbf{98.18\%}$ & $\textbf{88.45\%}$ & $\textbf{80.36\%}$  & $97.72\%$ & $94.51\%$ & $88.55\%$ & \textbf{91.48\%}\\
                DGI-HBOS & $96.02\%$ & $48.99\%$ & $0.00\%$ & $88.03\%$ & $46.82\%$ & $0.00\%$ &  47.91\% \\
                GraphSAGE-HBOS & $96.91\%$ & $68.39\%$ & $24.22\%$ & $\textbf{98.67\%}$ & $\textbf{96.69\%}$ & $\textbf{88.94\%}$ &  82.54\%\\

            \end{tabular}
            }
            \label{state_art_no_contamination}
        \end{table*}

         \begin{table*}[!ht]
            \renewcommand{\arraystretch}{1.3}
            \caption{Performance evaluation for metrics on both datasets compared with the baseline algorithms (4\% contamination).}
            \label{state_art_contamination}
            \centering
            \scalebox{0.8}{
           \begin{tabular}{c | *3c | *3c | *1c}
                \toprule
                \multicolumn{1}{c}{} &  \multicolumn{3}{c}{NF-UNSW-NB15-v2} & \multicolumn{3}{c}{NF-CSE-CIC-IDS2018-v2} &
               \multicolumn{1}{c}{Average Across Datasets}\\ 
                {} & Acc & Macro F1 & DR & Acc & Macro F1 & DR  & Macro F1 \\
                Anomal-E-PCA (Ours) & $\textbf{98.63\%}$ & $\textbf{92.18\%}$ & $\textbf{97.86\%}$ & $\textbf{97.11\%}$ & $\textbf{92.57\%}$ & $\textbf{79.16\%}$ & \textbf{92.38\%} \\

                DGI-PCA & $96.02\%$ & $48.89\%$ & $0.00\%$ & $88.03\%$ & $46.82\%$ & $0.00\%$ &  47.86\%  \\
                GraphSAGE-PCA & $95.95\%$ & $48.97\%$ & $0.00\%$ & $88.02\%$ & $46.82\%$ & $0.00\%$ & 47.90\%  \\

\hline
                 Anomal-E-IF (Ours) & $\textbf{98.66\%}$ & $\textbf{92.35\%}$ & $\textbf{98.77\%}$ & $\textbf{89.79\%}$ & $\textbf{81.11\%}$ & $\textbf{91.84\%}$ & \textbf{86.71\%} \\

                DGI-IF & $96.02\%$ & $48.99\%$ & $0.00\%$ & $88.03\%$ & $46.82\%$ & $0.00\%$ &  47.91\% \\
                GraphSAGE-IF & $60.55\%$ & $39.92\%$ & $24.57\%$ & $76.40\%$ & $66.56\%$ & $92.72\%$  &  53.24\% \\

\hline
                 Anomal-E-CBLOF (Ours) & $\textbf{98.57\%}$ & $\textbf{91.70\%}$ & $\textbf{95.72\%}$ & $97.80\%$ & $94.38\%$ & $\textbf{82.67\%}$ &  \textbf{93.04\%} \\
                DGI-CBLOF & $96.02\%$ & $48.99\%$ & $0.00\%$ & $88.03\%$ & $46.82\%$ & $0.00\%$ & 47.91\% \\
                GraphSAGE-CBLOF & $87.10\%$ & $49.53\%$ & $10.32\%$ & $\textbf{97.90\%}$ & $\textbf{94.61\%}$ & $82.60\%$  &  72.07\%\\
                
                \hline
                 Anomal-E-HBOS (Ours) &$\textbf{98.62\%}$ & $\textbf{91.89\%}$ & $\textbf{94.92\%}$   & $96.86\%$ & $91.89\%$ & $77.79\%$& \textbf{91.89\%}  \\
                DGI-HBOS & $96.02\%$ & $48.99\%$ & $0.00\%$ & $88.03\%$ & $46.82\%$ & $0.00\%$  & 47.91\%  \\
                GraphSAGE-HBOS & $88.60\%$ & $54.77\%$ & $26.63\%$ & $\textbf{97.90\%}$ & $\textbf{94.61\%}$ & $\textbf{82.60\%}$ & 74.69\%

                \\
            \end{tabular}
            }
        \end{table*}

\section{Conclusion}\label{conclusion}
To summarise, we presented Anomal-E, a self-supervised GNN that incorporates and leverages edge features for network intrusion and anomaly detection. Anomal-E can generalise and adapt to new, sophisticated attacks in both artificial and real network environments. To the best of our knowledge, this is the first approach to network intrusion detection utilising these factors for learning and evaluation. Through our experimental analysis and evaluation of the model, we established the benefit of using Anomal-E embeddings in attack identification. This was solidified by our comparison to the use of raw features on two benchmark NIDS datasets. Our experimental evaluation also displayed Anomal-E's ability to generalise to different datasets and clearly shows that the model performs exceptionally well compared to baseline methods. This further demonstrates the capabilities and potential of self-supervised GNNs for NIDSs. Moving forward, we plan to evaluate Anomal-E using real network traffic to gather informative insights into model performance outside of synthetic datasets. We also plan on investigating the effects of temporal-based training to further the exploration of Anomal-E's performance outside synthetic network conditions.

\bibliography{main}

\end{sloppypar}
\end{document}